# Space Expansion of Feature Selection for Designing more Accurate Error Predictors

Shayan Tabatabaei Nikkhah
Department of Electrical Engineering
Eindhoven University of Technology
Eindhoven, The Netherlands
s.tabatabaei.nikkhah@tue.nl

Mehdi Kamal
School of Electrical and Computer Engineering
University of Tehran
Tehran, Iran
mehdikamal@ut.ac.ir

Ali Afzali-Kusha
School of Electrical and Computer Engineering
University of Tehran
Tehran, Iran
afzali@ut.ac.ir

Massoud Pedram
Department of Electrical Engineering
University of Southern California
Los Angeles, CA, USA
pedram@usc.edu

## ABSTRACT

Approximate computing is being considered as a promising design paradigm to overcome the energy and performance challenges in computationally demanding applications. If the case where the accuracy can be configured, the quality level versus energy efficiency or delay also may be traded-off. For this technique to be used, one needs to make sure a satisfactory user experience. This requires employing error predictors to detect unacceptable approximation errors. In this work, we propose a scheduling-aware feature selection method which leverages the intermediate results of the hardware accelerator to improve the prediction accuracy. Additionally, it configures the error predictors according to the energy consumption and latency of the system. The approach enjoys the flexibility of the prediction time for a higher accuracy. The results on various benchmarks demonstrate significant improvements in the prediction accuracy compared to the prior works which used only the accelerator inputs for the prediction.

## 1 INTRODUCTION

The growing computational and performance demands of modern platforms have heralded a necessity of developing new techniques to unravel the challenges regarding energy and performance budgets. In this regard, one of the promising techniques, which has been vastly studied in recent years, is approximate computing which provides energy and performance gains at the cost of inaccuracies in outputs [1]. There are different parts in the emerging Recognition, Mining and Synthesis (RMS) applications which can be executed imprecisely to tradeoff quality for performance and energy gains [2].

Various approximate techniques, such as loop perforation [3], voltage scaling [4], and using approximate hardware accelerators [5], can be employed to execute the aforementioned error-resilient parts in an inexact manner. However, employing approximate computing methods may add some unexpected errors to final outputs. These errors strongly depend on application inputs [6], putting an extra emphasis on quality control in approximate computing. One of the proposed solutions for managing the output quality is sampling the output elements and changing the approximate configuration accordingly to meet the quality requirements [3], [7]. Given that output errors are strongly dependent to the inputs, this approach is highly susceptible to missing the output elements with large errors [6].

There is another approach emphasized in recent works (*e.g.,* [6], [8], [9]) which leverages light-weight quality checkers to dynamically manage the output quality. For each accelerator invocation, the light-weight checker (error predictor) decides whether the output error is greater than an error threshold or not, thereby issuing a rollback command to re-execute that iteration in a precise fashion. Three different error predictors are presented in [6] where two of them use accelerator inputs and one of them employs the accelerator outputs to do the prediction; likewise, other works in this scope (*e.g.,* [8]) use accelerator inputs for predicting errors. However, besides the accelerator inputs and outputs, the error predictor may employ intermediate results to predict the errors.

Having a high accuracy in both detecting the outputs with large errors and not issuing false rollback commands is of paramount importance in such systems. They are vital for meeting the quality demands and avoiding energy overheads due to unnecessary rollbacks respectively. Employing an ensemble of predictors is an approach to improve the accuracy of predictors [9]. However, this approach employs multiple predictors, which imposes hardware overheads in terms of area and energy consumption. These overheads may be unacceptable for relatively small designs which their energy budgets are tightly restricted. Additionally, manipulation of the dataset used for training the error predictor can be used to improve the prediction accuracy [10].

None of the prior works have used the intermediate results for predicting the approximation errors. Considering the fact that these intermediate results can combine the information of several accelerator inputs, they may have more information to separate the outputs with large errors from the ones with acceptable errors.

Hence, employing these intermediate results can improve the prediction accuracy. Furthermore, a smaller set of intermediate results may be used instead of the accelerator inputs set, which results in reducing the energy consumption and latency of prediction. Also, the prior works restricted the prediction time to the accelerator's latency. More timing flexibility, however, can improve the accuracy of prediction. Considering the large number of intermediate results of the accelerator, exhaustive search cannot be employed to select the most efficient subsets of these results due to its unacceptably large timing overhead (*e.g.,* several years for FFT benchmark). Moreover, these intermediate results become available (*i.e.,* usable) in different runtime cycles. Therefore, an effective approach is desired to select an efficient subsets of these results by considering their availability constraints without imposing a significant time overhead.

In this work, we argue that prediction accuracy can be significantly improved by selecting an appropriate subset of *features* from all the available ones (including accelerator inputs, intermediate results, and accelerator outputs). In this regard, we propose a *scheduling-aware feature selection* method which considers energy and time constraints (extracted from a scheduled hardware accelerator) to select the best configuration for error predictors (*e.g.,* number of features) as well as an efficient subset of features. In order to implement our method, we leverage the common feature selection techniques used in machine learning literature and modify them according to the constraints in our problem. Moreover, we offer the user flexibility in prolonging the prediction time, which can further improve the prediction accuracy. In sum, the contributions of our work include:

- Improving the prediction accuracy by selecting the most efficient features from all the available features in the accelerator.
- Analyzing the constraints for each application and choosing the best configuration for each predictor, instead of a fixed configuration for the predictors.
- Providing the option of prolonging the prediction time which offers a tradeoff between delay and prediction accuracy.

The rest of the paper is organized as follows. In Section 2, we review the related works and discuss our motivation. Our proposed method is elaborated in Section 3. The efficacy of our proposed method is assessed in Section 4. Finally, section 5 concludes the paper.

## 2 RELATED WORKS AND MOTIVATION

Managing the output quality is one of the key challenges in the recent approximate computing studies. Even though approximate computing can promise considerable energy and performance gains, quality control is essential for its practical uses. In order to control the average quality, SAGE [7] and Green [3] used a sampling approach which evaluated the quality once in *N* invocations and modified the intensity of approximation by comparing the quality with a pre-determined quality threshold. Although this sampling approach can be helpful to control the average quality, it lacks the ability to detect and prevent the occasional large errors.

Employing quality checkers is another approach for quality control. BRAINIAC [11] exploits application-specific light-weight checks to guarantee user-specified quality at runtime. This is performed by using less aggressive (*i.e.,* more accurate) accelerator configurations to meet the quality constraint. Rumba [6] presents three light-weight error predictors, namely linear predictor, decision tree, and moving average predictor, which predict whether the approximation error is greater than a given threshold or not. If so, rollback recovery (exact re-execution) is performed. Additionally, table-based predictors are proposed in [8] and [9]. These predictors generate a signature from accelerator inputs and map them to a table entry which stores a one bit label, indicating whether to rollback or not.

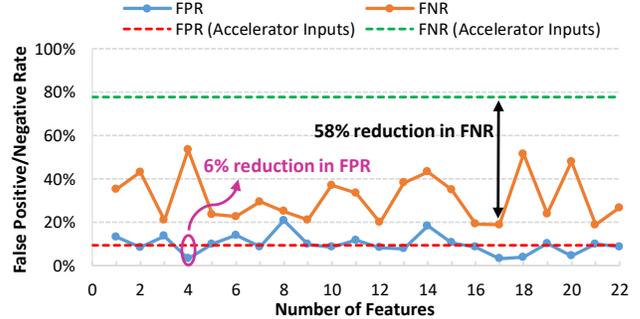

**Figure 1: False positive and false negative rates for various number of features in comparison with the case that accelerator inputs are used as features in Sobel benchmark.**

An iterative training algorithm is proposed in [10] which judiciously selects the training data and tunes the error threshold dynamically. In that work, neural-networks are exploited for both the accelerator and the error predictor. ApproxQA [12] presents a two-level controller, in which the configuration of the accelerator is determined by a high-level approximation controller at a coarse-grained scale, and rollback recoveries are decided by a rollback controller at a fine-grained scale. It employs the error predictors which are used in the prior works.

These studies demonstrate that employing the error predictors is a decent solution to the quality control problem. However, the error predictors do not perform perfectly. In order to improve the prediction accuracy, ensemble methods which employ several basic predictors are used in [9]. At the same time, this approach increases the energy consumption of prediction which might not be tolerable in some designs. Also, in order to cope with the complex behavior of applications, using a neural predictor is suggested in [10]. However, in this case, the energy overheads can overweight the energy gains in small designs. Therefore, we propose a feature selection approach which improves the prediction accuracy without further overheads.

None of the proposed error predictors in prior works employs intermediate results to predict the approximation error. While the predictor based on moving average in [6] uses accelerator outputs for prediction, all the other proposed predictors predict the error based on the accelerator inputs.

The example shown in Fig. 1 demonstrates how employing intermediate results as features can improve the accuracy of prediction. In this example, an accelerator which employs approximate arithmetic units is used to execute the convolution loop in *Sobel* benchmark and a decision tree with depth of 4 is used to predict errors. The error threshold is determined in such a way that rollback recoveries are needed for 10% of the invocations. The dashed lines show false positive and false negative rates (FPR and FNR) for the case in which accelerator inputs (9 features) are used to predict errors, while the solid lines correspond to the case where different number of features are *randomly* selected from all the features (including accelerator inputs, intermediate results, and accelerator outputs). This study shows that FNR and FPR can be considerably reduced by selecting proper features (up to 58%). Moreover, it can be seen that FPR and FNR values do not change in the same way for different number of features. Hence, an appropriate metric (merit function) is needed to select the best subset of features. As we will discuss in the next sections, we employ *F1 score*, which considers both the false positives and the false negatives, to evaluate our feature subsets. Moreover, a proper algorithm is needed to find the best feature subset. In Section 3.3, we will discuss different feature selection algorithms.

# 3 PROPOSED APPROACH

## 3.1 Overview

In this work, we apply our approach to a platform comprised of a host, an accelerator, and an error predictor which sends rollback commands (re-execution commands) to the host to re-execute the application. Our proposed feature selection flow is depicted in Fig. 2. The user provides the scheduled dataflow graph of the accelerator, prediction time and energy constraints, as well as training and test datasets. These inputs are employed to, first, configure the error predictor and, then, do the feature selection.

Generally, designers prefer to restrict the prediction time to the time which the accelerator needs to compute the outputs. In this case, however, some features (such as the outputs which are generated in the last cycle of accelerator invocation) cannot be selected. Accordingly, in the proposed method, we let the designer specify whether prediction results are allowed to be generated after accelerator results or not by defining *prediction time* constraint.

Each predictor has specific parameters (configuration) which must be specified before the feature selection. This configuration (*e.g.,* depth of the decision tree and number of features in the linear predictor) is determined according to the energy and time constraints. Additionally, in order to employ the intermediate results without violating the prediction time constraint (*e.g.,* the latest cycle at which prediction result must be ready), the availability of each feature in each operating cycle should be determined. The cycle times can be extracted from the scheduled DFG. With all this information, feature selection can be performed. Now, in the following sections, we first elaborate how to determine the configuration of the error predictors, then we discuss feature selection algorithms and the way our constraints affect them.

## 3.2 Configuring the Error Predictors

Each error predictor has specific configurations which can be determined based on the prediction time and energy constraints of the system. In this work, we narrow our study to the linear predictor and decision tree, which are light-weight enough for most of the applications. However, other types of predictors can be analyzed in the same way (*e.g.,* the number of neurons and hidden layers in neural error predictor).

*3.2.1 Linear Predictor.* The linear predictor is comprised of a MAC unit and storage elements [6]. This predictor computes a linear function of features using their values and pre-determined coefficients. Accordingly, the key parameter (from feature selection perspective) to be determined is the number of features. Considering the prediction time constraint, the upper bound for the number of features in the linear predictor ($FL_{UB,\tau}$) is obtained by

$$FL_{UB,\tau} = \left\lfloor \frac{\tau_{const}}{T_{MAC}} \right\rfloor \quad (1)$$

where $\tau_{const}$ is the prediction time constraint and $T_{MAC}$ is the latency of one MAC operation (including the time needed for reading coefficients from buffers).

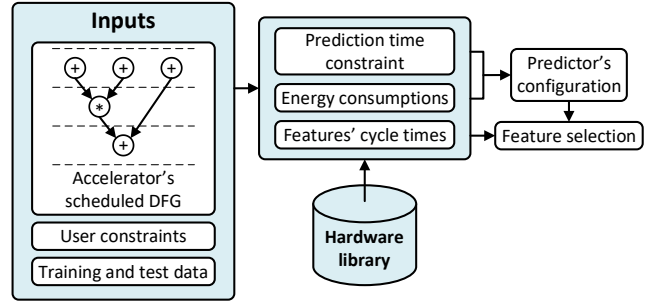

Figure 2: The flow of the proposed approach.

However, energy budget can also limit the number of features. The energy constraint can be described as:

$$E_{accelerator} + R \times E_{rollback} + E_{predictor} \leq \alpha \times E_{exact} \quad (2)$$

where $E_{accelerator}$ is the energy consumed by the approximate accelerator in one invocation, $R$ is the rate of rollback recoveries which can be estimated using the training dataset and the error threshold, $E_{rollback}$ is the energy cost of one re-execution, $E_{predictor}$ is the energy consumption of one prediction, $\alpha$ is the expected energy gain, and $E_{exact}$ is the energy consumption of one exact (by employing host or an exact accelerator) invocation. Typically, $E_{rollback}$ and $E_{exact}$ are equal. According to this equation, the average energy consumption of the system (including the energy consumption of the accelerator, the error predictor, and the cost of rollbacks) must be lower than $\alpha$ times of the average energy consumption of the case that no approximations are used. Given the fact that $E_{accelerator}$, $E_{rollback}$, $E_{exact}$, and $R$ (approximately) are available through estimation, we can calculate an upper bound for $E_{predictor}$ for a desired energy gain of $\alpha$. By doing so, another upper bound for the number of features in the linear predictor ($FL_{UB,E}$) can be obtained as:

$$FL_{UB,E} = \left\lfloor \frac{E_{predictor}}{E_{MAC}} \right\rfloor \quad (3)$$

where $E_{MAC}$ is the energy consumption of one MAC operation (including the energy of accessing the buffers). Finally, the upper bound for the number of features for a linear predictor ($FL_{UB}$) can be obtained from $FL_{UB} = Min(FL_{UB,\tau}, FL_{UB,E})$.

*3.2.2 Decision Tree.* The decision tree is made up of a comparator, a controller, and storage elements [6]. The number of comparisons required to predict the error corresponds to the depth of the tree. Hence, the target parameter to be specified in this predictor is its depth. Similarly, two upper bounds can be determined for the depth of the decision tree ($FD_{UB,\tau}$ and $FD_{UB,E}$). The first one is obtained using (1) where $T_{traverse}$, which is the number of cycles spent to traverse one level in the tree, is used instead of $T_{MAC}$. The second upper bound is determined similar to $FL_{UB,E}$ in (3) where $E_{traverse}$ is used instead of MAC energy, and the depth of the tree is determined by $FD_{UB} = Min(FD_{UB,\tau}, FD_{UB,E})$.

## 3.3 Common Feature Selection Algorithms

Feature selection is vastly used in the machine learning literature to reduce irrelevant and redundant data, thereby improving the performance (*e.g.*, accuracy) of classification [13]. The aim is to eliminate the data which has no extra information in separating different classes. Besides improving the prediction accuracy, feature selection may reduce the prediction time by decreasing the number of features. Although feature selection can be performed through an exhaustive search, the search time grows exponentially with increasing the number of features. For instance, for an accelerator with $n$ inputs, $m$ intermediate results, and $k$ outputs, the total number of feature subsets is $2^{m+n+k}$, which makes the exhaustive search unfeasible for most of the designs. Therefore, we leverage the most popular feature selection algorithms in our approach which are discussed in following subsections.

*3.3.1 Filter Methods.* These algorithms sort the features according to a merit value (*e.g.*, Pearson correlation coefficient) and eliminate the features whose merit values are lower than a pre-defined threshold. Given that the filter methods act independently of learning algorithms, they are relatively fast. In this work, we employ variance, Pearson correlation coefficient, and Fisher discriminant ratio as the merits for the filter methods.

*3.3.2 Wrapper Methods.* In these algorithms, heuristic search methods are employed to find a sub-optimal feature subset, and a learning algorithm (a predictor) is used to evaluate feature subsets. Since these methods need to train and test predictors in their search flow, they are slower than the filter methods. On the other hand, feature subsets are evaluated in practice; hence, they are more likely to find a better subset. In this work, the following heuristics are considered:

- **Sequential Forward Selection:** Starting from no features, this algorithm selects one feature in each iteration (according to a merit value) and continues until the target number of features is selected.
- **Sequential Backward Elimination:** In each iteration, the algorithm eliminates one feature and continues until the number of selected features is reduced to the target number.
- **Particle Swarm Optimization (PSO):** A population based stochastic optimization technique which starts with a population of random solutions, and updates them in each algorithm iteration.
- **Quantum Genetic Algorithm (QGA):** A genetic algorithm based on the principle of quantum computation which has faster convergence speed than the traditional genetic algorithm.

In order to evaluate the feature subsets while using the wrapper methods, we suggest employing the *$F_1$ score*, which is the harmonic mean of precision and recall:

$$F_1 = 2 \times \frac{precision \times recall}{precision + recall} \quad (4)$$

where *precision* is the ratio of true positives to total number of predicted positives (sum of true positives and false positives), and *recall* is the ratio of true positives to total number of real positives (sum of true positives and false negatives, referred to as coverage of large errors in [6]). *Precision* shows what percentage of our rollback commands are correct commands. A low *precision* corresponds to numerous unwanted rollbacks, which imposes energy overheads. *Recall* demonstrates how successful we are in detecting the large errors. Accordingly, *$F_1$ score* considers both the quality and the energy.

## 3.4 Scheduling-Aware Feature Selection

As mentioned before, we employ intermediate results to improve the prediction accuracy. However, these results are not available at the beginning of each accelerator invocation and may become available after several cycles. Accordingly, we formulate the availability of the features by defining *feature-availability* constraint, and modify the aforementioned feature selection algorithms (Section 3.3) by adding this constraint to their selection rules. For the linear predictor, this constraint is defined for each accelerator invocation cycle time, and for the decision tree, it is described for each feature. As mentioned in previous sections, we narrow our focus on these two predictors in this work, and other types of predictors can be studied likewise.

*3.4.1 Feature-availability Constraint.* In the following, this constraint is discussed for the linear predictor and decision tree separately.

**Linear Predictor:** In the case of the linear predictor, we define this constraint (based on (1) and (3)) for $i^{th}$ cycle of the accelerator invocation ($AFL_i$, $i \in [1, \lambda]$ where $\lambda$ is latency of the accelerator) as:

$$AFL_i = min\left(\left\lfloor \frac{\tau_{const} - i + 1}{T_{MAC}} \right\rfloor, FL_{UB,E}\right) \quad (5)$$

where $\tau_{const} - i + 1$ shows the available prediction time in $i^{th}$ cycle, and $AFL_i$ specifies the maximum number of features that can be selected from the ones becoming available in the $i^{th}$ cycle of each accelerator invocation. Note that for the accelerator inputs, $i$ equals to one.

**Decision Tree:** The *feature-availability* constraint for the decision tree shows the minimum possible position in the tree that a feature is allowed to be selected by the training algorithm to perform node split, and is defined as:

$$AFL_j = \left\lceil \frac{i + (T_{traverse} \times D) - \tau_{const} - 1}{T_{traverse}} \right\rceil \quad (6)$$

where $i$ is the cycle index where $j^{th}$ feature becomes available, $D$ is the depth of tree (determined in Section 3.2.2), $\tau_{const}$ is the latency of prediction, and $AFL_j$ is the minimum position in the tree that $j^{th}$ feature can be selected. If the value obtained from (6) is negative, zero is considered for it, which corresponds to the root node in the tree.

*3.4.2 Feature Selection.* Employing the *feature-availability* constraint, we modify the aforementioned feature selection algorithms which are discussed in the following paragraphs.

**Filter Methods:** For the linear predictor, after the features are sorted according to their merit values (*e.g.*, variance values), the algorithm starts from the top of the list and selects one feature at a time. For the candidate feature which becomes available in the $i^{th}$ cycle of accelerator invocation, the modified algorithm (considered the feature-availability constraint) examines whether the number of features currently selected from the ones which are available after the $i^{th}$ cycle is lower than $AFL_i$ or not. If so, the feature will be selected; otherwise, it will be discarded, and next features on the list will be assessed. These steps are repeated until the target number of features are selected.

In the decision tree, after applying the filter methods, the training algorithm selects the best feature to perform the node split for each node. Accordingly, we modify the training algorithm by applying the *feature-availability* constraint stated in (6). For this purpose, when the training algorithm selects $j^{th}$ feature for the node split in the $p^{th}$ position of the tree, we propose to examine whether $AFL_j$ (the minimum possible position in the

Table 1: Benchmarks and their details.

| Benchmark | Domain | Train/Test Data | Inputs | Total Features | Linear Predictor Features | Tree Depth |
|---|---|---|---|---|---|---|
| Sobel | Image Processing | 5K pixels | 9 | 22 | 4 | 4 |
| FIR Filter (6-tap) | Signal Processing | 5K random samples | 6 | 17 | 6 | 6 |
| Forwardk2j | Robotics | 5K random (x,y) angles | 2 | 27 | 4 | 4 |
| DCT (8-point) | Compression | 5K pixels | 8 | 50 | 11 | 7 |
| FFT (8-point) | Signal Processing | 5K random samples | 8 | 48 | 14 | 7 |

**Algorithm 1:** Sequential Forward Selection

```
1   sel_features ← [] // selected features
2   cyc_features ← zeros(λ) // number of selected features from
    each cycle
3   while len(sel_features) < target number of features do
4       merits ← [] // merit values
5       for each f in features do
6           append f to sel_features
7           if predictor = Tree then
8               merit ← modified_tree_training
9           else
10              merit ← linear_training
11          end
12          append merit to merits & remove f from
            sel_features
13      end
14      Sort (merits) // sort from greatest to least
15      if predictor = Tree then
16          append corresponding feature of merits[0] to
            sel_features
17      else
18          selected ← 0 , i ← 0
19          while selected = 0 & i < len(merits) do
20              f ← corresponding feature of merits[i]
21              if cyc_features[f.cycle] < M [f.cycle] then
22                  append f to sel_features , selected ← 1
23                  cyc_features[f.cycle] ++
24              else
25                  i ++
26              end
27          end
28          break if selected = 0 // no more legit features
29      end
30  end
```

tree) is greater than $p$ or not. If so, the $j^{th}$ feature will not be selected for the node split, and other features will be assessed.

**Wrapper Methods:** In these methods, according to their heuristic algorithm, several subsets of features are determined and evaluated in each step where one or more features are selected or eliminated. To select (eliminate) each feature, same as the filter methods, our approach employs the *feature-availability* constraint for the linear predictor in selection (elimination) phase, and for the decision tree in the training phase.

*3.4.3 Modified Sequential Forward Selection: An Example.* For a better understanding, as an example, the details of the modified sequential forward selection algorithm is shown in Algorithm 1. Initially, no features are selected; hence, *sel_features* is an empty array, and *cyc_features*, which shows the number of features selected from different cycles, is an array with zero elements. Then, in each step, the algorithm selects one feature (line 3-30). To do so, the influence of adding one feature is assessed by adding one feature at a time (line 6) and obtaining a merit value using training and testing a predictor (line 7-12). Afterwards, it sorts the merit values from greatest to least (line 14). As mentioned before, we embed the feature-availability constraint of the features into the training algorithm of the trees using (6). Therefore, if the predictor is decision tree, the feature with the highest merit is selected, and applying the feature-availability constraint is handled in the training algorithm (line 15-16).

However, if the predictor is from the linear type, the algorithm starts from the top of the list (sorted according to their merit values) and examines whether the number of features which are currently selected from cycle time of the candidate feature is lower than the maximum allowed value ($AFL_i$ constraints) or not (line 21). If true, the candidate feature will be selected, and the corresponding cycle time in *cyc_features* will be incremented by one (line 22-23). In other case, the next feature in the list will be the new candidate. If none of the features satisfies the constraint, the algorithm will be aborted by a break statement (line 28).

## 4 RESULTS AND DISCUSSION

### 4.1 Experimental Setup

To assess the efficacy of the proposed approach, five benchmarks, including Sobel, FIR filter, Forwardk2j, DCT, and FFT, have been considered in the studies which their details are presented in Table 1. In order to build the approximate accelerator, the approximate arithmetic units proposed in [14], [15] and the scheduling algorithm presented in [16] have been employed. The error predictors have been designed using exact carry-look ahead adders and Dadda multipliers. The designed accelerators and predictors have been described by Verilog HDL and synthesized by Synopsys Design Compiler using NanGate 45nm library [17]. The energy consumptions were extracted using Synopsys Power Compiler and CACTI 6.5 [18].

### 4.2 Experimental Results

*4.2.1 Run Time Comparison.* The average time taken to perform the feature selection under different modified feature selection algorithms are shown in Fig. 3a. In this study, four different scenarios where the rollback ratios are 20%, 15%, 10%, and 5% have been considered. This study has been performed on a system with a Core i7 6700HQ CPU and 8 GBs of RAM. Var, FDR, and PCC correspond to the filter methods which use feature variances, Fisher discriminant ratios, and Pearson correlation coefficients, respectively. SFS and SBE denote sequential forward selection and sequential backward elimination, respectively. PSO and QGA algorithms were executed with swarm size of 20 and 20 chromosomes, respectively, and for the both algorithms 50 iterations has been considered as the termination conditions. As the reported run times show, the filter methods are significantly faster than the wrapper methods. Generally, in the studied cases, their run times were less than 0.5 seconds. It is worth noting that the exhaustive search for finding the best feature subsets in FFT benchmark will take 360 years (based on our estimations), which puts an emphasis on employing feature selection algorithms.

*4.2.2 Prediction Accuracy Comparison.* As mentioned in Section 3.3, we used $F_1$ score to evaluate the prediction accuracy. In Fig. 3b and 3c, the scores of different feature selection algorithms are demonstrated for the scenario of 10% rollback rate. Moreover, they are compared with the case that only the accelerator inputs were employed for prediction. It can be seen that in all the studied cases, employing the intermediate results improves the prediction accuracy (2.3x and 24.5x average improvements in $F_1$ score for the decision tree and the linear predictor respectively). The very low $F_1$ scores (*e.g.,* below 0.01) for the linear predictor were caused by extreme tendency of this predictor to the zero class (*i.e.,* small errors class). In this case, *recall* value converged to zero, resulting in a very low $F_1$ score. Furthermore, the results show that the decision tree outperforms the linear predictor in all the studied benchmarks, which is proximately in line with Rumba's findings [6].

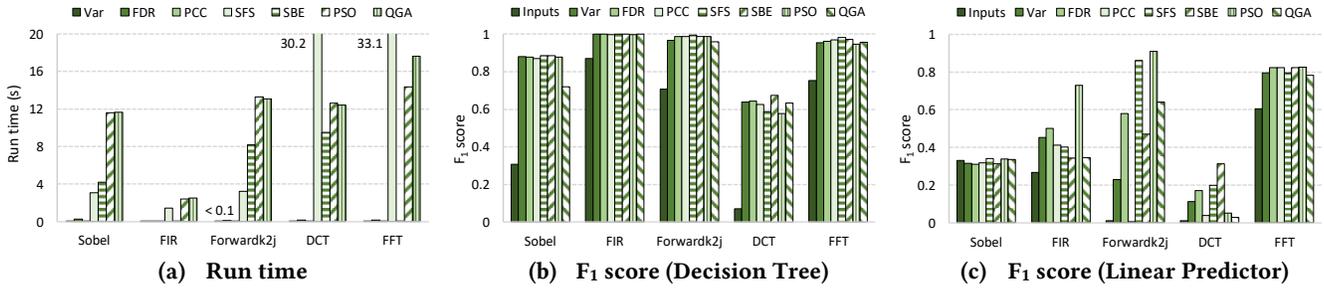

**Figure 3: Comparison of feature selection algorithms.**

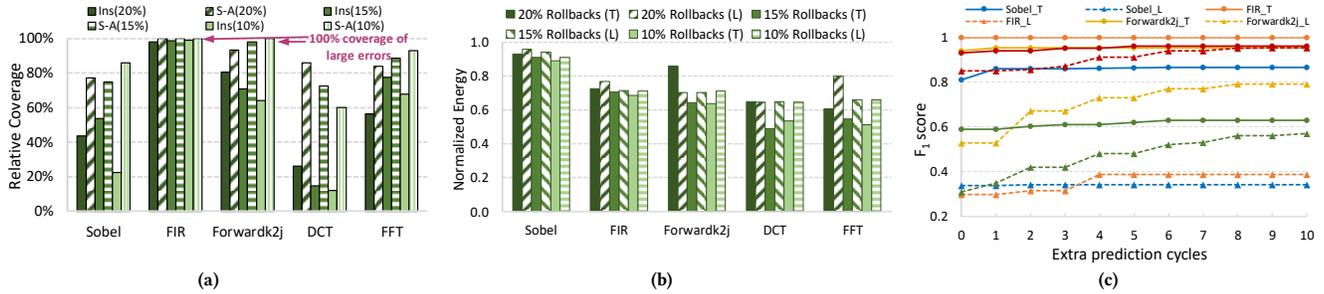

**Figure 4: a) The relative coverage of large errors for the decision tree under different rollback ratios, b) The normalized energy of the system under different rollback ratios, c) The influence of extra prediction time on prediction accuracy.**

*4.2.3 Coverage of Large Errors.* Although we have shown that our feature selection method has considerably improved the $F_1$ score, it is important to study the influence of our approach on the coverage of large errors (also known as recall). In Fig. 4a, the relative coverage of large errors for the case of using decision tree under three rollback ratios are shown. The solid bars correspond to the case that only accelerator inputs were used for feature selection, while the hatched bars depict the results of the proposed scheduling-aware (S-A) feature selection method. It can be seen that in all the studied cases, the proposed approach has improved the coverage of large errors (from 1% in FIR to 64% in Sobel benchmark). Moreover, in FIR and Forwark2j benchmarks, the coverage of 100% was achieved by employing the proposed approach.

*4.2.4 Energy Consumption.* In order to evaluate the total energy consumption (including the energy consumed by the approximate accelerator, error predictor, and rollback recoveries) all the values are normalized to the case where an exact accelerator (without error predictor) is employed to execute the benchmarks. For this part, an expected energy gain (α in (2)) of 0.7 was considered. As mentioned in Section 3.2, the maximum energy consumption of error predictor is determined in such a way that the total energy consumption of the system becomes lower than α times of total energy consumption of the case that no approximations are used. Therefore, in all the experiments, we expect that the normalized energy becomes lower than α. In Fig. 4b, the normalized energy consumption of system under three different rollback ratios is shown. In this figure, solid bars correspond to the experiments where decision tree is used, while the hatched bars correspond to the ones in which linear predictor is employed. According to Fig. 4b, in all the cases, the normalized energy is lower than 1.0, which shows that employing approximate computing has reduced the energy consumption. Additionally, in most of the cases, the normalized energy is lower than 0.7, which meets our expected energy gain. However, in Sobel benchmark and few cases in other benchmarks, the normalized energy is larger than the expected energy gain (up to 26%, and 12% on average) due to unnecessary rollbacks (False Positives) which are caused by imperfections of error predictors.

*4.2.5 Influence of Prediction Time.* As mentioned in the previous sections, the proposed method let the user specify extra cycles for the prediction time. These extra cycles provide the opportunity for the feature selection algorithms to employ more features from the latest cycle times. The influence of these extra cycles is depicted in Fig. 4c for both the decision tree (solid lines) and the linear predictor (dashed lines). It is demonstrated that $F_1$ scores have increased by increasing the number of extra cycles in all the studied cases. However, the score growth in the linear predictor is more considerable. The $F_1$ scores have been increased by 3.6% and 35.1% on average in the decision tree and the linear predictor respectively. Although the extra prediction time leads to exploiting more features by the predictor, the number of features are restricted by the energy constraints, and using more features in the decision tree needs a deeper tree and bigger memories to store the coefficients, which significantly increases the total energy consumption. On the other hand, the extra overhead is not that intense in the linear predictor. Therefore, as it can be seen in

Fig. 4c, the scores have been increased significantly for the linear predictor (up to 83% in DCT benchmark).

## 5 CONCLUSIONS

In this work, we presented a feature selection approach which utilized the intermediate results of a hardware accelerator in a heterogeneous system (comprised of a host, an accelerator, and an error predictor) to improve the prediction accuracy. It also configured the error predictors according to the constraints of the system. Moreover, it provided a delay-accuracy tradeoff for the error predictors. Our studies on various benchmarks showed substantial improvements in the prediction accuracy compared to the prior works.